# On the Consistency of Fairness Measurement Methods for Regression Tasks

**Abdalwahab Almajed**[1] , **Maryam Tabar**[2] , **Peyman Najafirad**[2]
[1]Imam Abdulrahman Bin Faisal University
[2]The University of Texas at San Antonio
analmajed@iau.edu.sa, {maryam.tabar, peyman.najafirad}@utsa.edu

## Abstract

With growing applications of Machine Learning (ML) techniques in the real world, it is highly important to ensure that these models work in an equitable manner. One main step in ensuring fairness is to effectively measure fairness, and to this end, various metrics have been proposed in the past literature. While the computation of those metrics are straightforward in the classification setup, it is computationally intractable in the regression domain. To address the challenge of computational intractability, past literature proposed various methods to approximate such metrics. However, they did not verify the extent to which the output of such approximation algorithms are consistent with each other. To fill this gap, this paper comprehensively studies the consistency of the output of various fairness measurement methods through conducting an extensive set of experiments on various regression tasks. As a result, it finds that while some fairness measurement approaches show strong consistency across various regression tasks, certain methods show a relatively poor consistency in certain regression tasks. This, in turn, calls for a more principled approach for measuring fairness in the regression domain. The codes can be accessed via https://github.com/wahab1412/Fairness-Measurement-ML.

## 1 Introduction

With recent advances in ML, there has been growing interests in employing ML techniques in decision-making tasks in the field. Although this transition holds the promise of enhanced efficiency and ostensibly unbiased decisions, empirical findings sometimes reveal inconsistencies with these expectations [Mehrabi *et al.*, 2021]. An example of bias has been identified in ML hiring model; i.e., Upon further examination, Amazon discovered that the model exhibited a bias against female applicants, preferentially selecting male candidates for software developer and technical positions [Carnegie Mellon University, 2018]. Another instance in healthcare revealed that a model assigned a higher 30-day readmission rate to Black patients compared to non-Black patients [Wang *et al.*, 2024]. Therefore, it is highly important to ensure fair function of ML models so that its deployment does not exacerbate inequities in society.

A fundamental step in ensuring fairness is to accurately and reliably measure that. Unfortunately, defining fairness and its metrics is complex due to the absence of a universally accepted definition. Prior work proposed various individual fairness and group fairness definitions and metrics [Caton and Haas, 2020]. However, focusing on the classification tasks, most of those fairness measurement algorithms are designed for situations where the target variable is a categorical variable, and those metrics are computationally intractable in regression domain where the target variable is a continuous numeric variable. To address this challenge, past literature introduced methods to approximate (not exactly measure) such metrics and tried to address that from various perspectives [Caton and Haas, 2020]. For example, recent research [Steinberg *et al.*, 2020; Baniecki *et al.*, 2021] relied on a density ratio estimation approach to approximate the demographic parity metric. Additionally, Agarwal et al. [2019] proposed to reduce regression task into a classification task with a large number of classes to facilitate fairness computation in regression. However, to the best of our knowledge, there is no prior work investigating whether various approximation methods that measures the same fairness metric show consistent outcome; i.e., Given fairness metric $f$ and two ML regression models $m_1$ and $m_2$, if method $a_1$ shows that $m_1$ is more fair than $m_2$ in terms of $f$, then method $a_2$ shows the same trend.

To fill this research gap, our paper aims to answer the following research question: *Do various algorithms approximating the same fairness metric in regression show consistent outcome?* To address this question, we focused on two main fairness metrics: Parity-based metric (which measures the extent to which a model's predictions are independent of belonging to a protected group), and confusion matrix-based metric (which checks if a model is performing equally well across various groups) [Caton and Haas, 2020]. Then, to study the consistency of the outcome of methods approximating these metrics in regression domain, we take the following steps: 1) we train an extensive set of ML models on a regression task, 2) we measure the fairness of their outcome using various approximation methods, and finally 3) we compute the correlation between the output of various approximation methods using the Spearman and Pearson correlations.

The results indicate significant variability in the consistency of the outcome of fairness measurement methods across various tasks. For example, among methods approximating parity-based fairness metric, Demographic Parity through Reduction to Classification (P1) [Agarwal *et al.*, 2019] and Demographic Parity with Wasserstein Barycenters (P2) [Chzhen *et al.*, 2020] show a relatively high consistency across all considered regression tasks, however, P1 and Independence via Probabilistic Classifier-based Density Ratio Estimation (P4) [Steinberg *et al.*, 2020] show a poor alignment on certain regression tasks. Similarly, in the case of methods approximating the confusion matrix-based metric, the methods proposed in [Steinberg *et al.*, 2020] and [Mary *et al.*, 2019] do not seem to align well in most cases.

The discovered inconsistency is very important because unreliability of fairness measurement methods could result in misinterpretation of the equity of ML model's performance, and also, misinterpretation of the effectiveness of various bias mitigation algorithms in the regression domain. Therefore, we advise researchers to utilize multiple fairness metrics to improve both the reliability and the interpretability of fairness evaluations. These observations also call for fundamental research in the fair regression domain to develop a reliable method for measuring various fairness metrics.

## 2 Literature Review

As ML applications in decision-making proliferate, it becomes imperative to ensure that these models operate fairly and are devoid of biases. One main step in ensuring fairness is to accurately and reliably measure the amount of biases, and to this end, parity-based and confusion matrix-based metrics [Caton and Haas, 2020] are widely-used metrics defined as follows: Suppose that $A$, $Y$, and $S$ refer to the protected attribute (such as race), ground-truth target variable, and the ML model's prediction, respectively. Parity-based measures the extent to $S$ and $A$ are statistically independent (i.e., $S \perp\!\!\!\perp A$), and confusion matrix-based metric measures the extent to $S$ and $A$ are statistically independent given $Y$ (i.e., $S \perp\!\!\!\perp A|Y$) [Steinberg *et al.*, 2020]. While the computation of these metrics is straightforward in classification domain where the target variable is categorical, it is computationally intractable in regression domain where the target variable is a continuous numeric variable.

To quantify biases in regression tasks, various solutions have been proposed to approximate parity-based and/or confusion matrix-based metrics in regression. For example, Steinberg et al. [2020] reduces this problem into density-ratio estimation problem, and leveraged probabilistic classifier-based approaches to estimate those metrics. Additionally, to address the challenges posed by continuous numeric variable, Agarwal et al. [2019] reduces the fair regression problem into a classification problem and discretize the target variable into a large number of classes. Further, Mary et al. [2019] examined the measurement of parity-based metric through the Hirschfeld-Gebelein-Rényi (HGR) Maximum Correlation Coefficient, and chzhen et al. [2020] utilizes the Kolmogorov-Smirnov distance to compare the cumulative distribution functions (CDFs) across various groups, and facilitate the measurement of fairness in regression.

While there have been various attempts to approximate fairness metrics in regression, to the best of our knowledge, there is no prior work investigating the consistency of the outcome of these methods. This paper aims to fill this gap through a comprehensive set of experiments.

## 3 Methodology

This paper investigates the consistency of various methods proposed for measuring parity-based and confusion matrix-based fairness metrics in the regression domain. To this end, we take the following steps: (1) for each considered regression task, we develop a wide range of predictive ML models, (2) we compute the fairness of their predicted values in terms of parity-based and confusion matrix-based metrics using various existing approximation methods, and finally, (3) for each metric, we compute the correlation between the fairness measures obtained from each pair of approximation methods using Pearson and Spearman correlations and analyze if there is a high linear/monotonic association between the output of various measurement algorithms[1]. The following paragraphs elaborate on each step.

As mentioned before, in this study, we focus on parity-based fairness metric and confusion matrix-based fairness metric. For the parity-based metric, we study the consistency of the outcome of the following four methods: (P1) Demographic Parity through Reduction to Classification [Agarwal *et al.*, 2019], (P2) Demographic Parity with Wasserstein Barycenters [Chzhen *et al.*, 2020], (P3) Demographic Parity via Rényi Correlation [Mary *et al.*, 2019] and (P4) Independence via Probabilistic Classifier-based Density Ratio Estimation [Steinberg *et al.*, 2020]. For the confusion matrix-based metric, we study the consistency of two methods, called (C1) Separation via Probabilistic Classifier-based Density Ratio Estimation [Steinberg *et al.*, 2020], and (C2) Equalized Odds [Mary *et al.*, 2019].

As for regression tasks, to comprehensively study the consistency in regression domain, we conduct experiments on three different regression tasks on three publicly available datasets: (1) The first task is to predict student GPA using the Law School dataset [Wightman, 1998], (2) The second one is to predict the percentage of violent crime using the Communities and Crime dataset [Redmond, 2009], and (3) the third task is to predict the insurance cost using the Insurance data [Lantz, 2019]. In the first two tasks, race is considered as the protected attribute, and for the third task, gender is considered as the protected attribute.

As for the underlying ML models, we developed 22 ML models for each regression task. This set of ML models includes both classical ML models (such as XGBoost [Chen and Guestrin, 2016], Random Forest [Breiman, 2001], and Support-Vector Machine [Cortes and Vapnik, 1995]) as well as neural network models (such as TabNet [Arik and Pfister, 2021] and Multi-Layer Perceptron). Detailed information about ML models and hyper-parameters are available in our code base at https://github.com/wahab1412/Fairness-Measurement-ML.

---

[1] Pearson and Spearman correlations ($\in [-1, 1]$) measure the extent to which the outcome of two methods are linearly correlated and monotonically correlated (respectively).

Table 1: Pearson and Spearman correlations between pairs of methods approximating the parity-based fairness metric (* shows the statistical significance with p-value of 0.05).

| Corr. Metric | Fairness Measurement | P1 | P2 | P3 | P4 |
|---|---|---|---|---|---|
| Pearson | P1 | - | (0.99*, 0.99*, 0.91*) | (0.96*, 0.80*, 0.71*) | (0.88*, 0.54*, 0.38) |
|  | P2 | (0.99*, 0.99*, 0.91*) | - | (0.96*, 0.79*, 0.75*) | (0.88*, 0.53*, 0.47*) |
|  | P3 | (0.96*, 0.80*, 0.71*) | (0.96*, 0.79*, 0.75*) | - | (0.88*, 0.62*, 0.25) |
|  | P4 | (0.88*, 0.54*, 0.38) | (0.88*, 0.53*, 0.47*) | (0.88*, 0.62*, 0.25) | - |
| Spearman | P1 | - | (0.99*, 0.99*, 0.90*) | (0.96*, 0.67*, 0.62*) | (0.95*, 0.83*, 0.42*) |
|  | P2 | (0.99*, 0.99*, 0.90*) | - | (0.96*, 0.68*, 0.57*) | (0.95*, 0.81*, 0.44*) |
|  | P3 | (0.96*, 0.67*, 0.62*) | (0.96*, 0.68*, 0.57*) | - | (0.97*, 0.71*, 0.22) |
|  | P4 | (0.95*, 0.83*, 0.42*) | (0.95*, 0.81*, 0.44*) | (0.97*, 0.71*, 0.22) | - |

## 4 Experimental Results

In this section, we provide empirical results on the consistency of the outcome of various methods approximating parity-based and confusion matrix-based metrics. Table 1 shows the obtained Spearman and Pearson correlation values across each pair of methods approximating the parity-based fairness metric. The cell corresponding to row $x$ and column $y$ shows the obtained correlation between the outcome of approximation methods $x$ and $y$ on the the Law School, Communities and Crime, and Insurance datasets, respectively.

According the results, there is a varying level of consistency between the outcome of various fairness measurement methods, and it changes across various regression tasks. In fact, the correlation between P1 and P2 is notably high across all datasets, with Pearson and Spearman values of consistently greater than/equal to 0.9. However, while P4 method shows a relatively high correlation with other methods on the Law School dataset (with average Pearson of 0.88 and Spearman of 0.95), it shows a poor consistency with other methods on the Insurance dataset (with an average Pearson and Spearman correlations of about 0.36). This shows that, depending on the data/regression task, these methods could show a varying level of consistency in approximating the parity-based fairness metric.

Additionally, we studied the consistency of the outcome of two methods (i.e., Separation via Probabilistic Classifier-based Density Ratio Estimation and Equalized Odds) measuring the confusion matrix-based fairness metric and Table 2 shows the results. According to the results, these two methods shows a variable level of consistency across various datasets, and similar to before, they show a poor consistency on the Insurance dataset. These observations confirm that the possibility of high inconsistency between the output of various methods, and therefore, further research needs to be conducted to address the challenge of fairness measurement in regression tasks.

We, now, dig deeper into the output of various approximation methods to better analyze the consistency. To this end, we visualize the fairness values derived from each pair of methods using scatter plots. Figure 1 represents the fairness values derived from (C1 and C2) for the 22 developed ML models across three datasets. In this figure, x-axis and y-axis correspond to different fairness measurement methods, the fairness of each ML model is shown with a dot, and the result on different datasets is shown with a different color. A key observation is the anomalies that can be seen in distinct regions for the Law School and Communities and Crime datasets; i.e., sometimes, while one approximation method suggests that the confusion matrix-based disparity is increasing, the other method suggests that the disparity is decreasing.

Table 2: Pearson and Spearman correlations between methods approximating the confusion matrix-based fairness metric.

| Corr. Metric | |
|---|---|
| Pearson | (0.83*, 0.50*, -0.31) |
| Spearman | (0.86*, 0.64*, -0.17) |

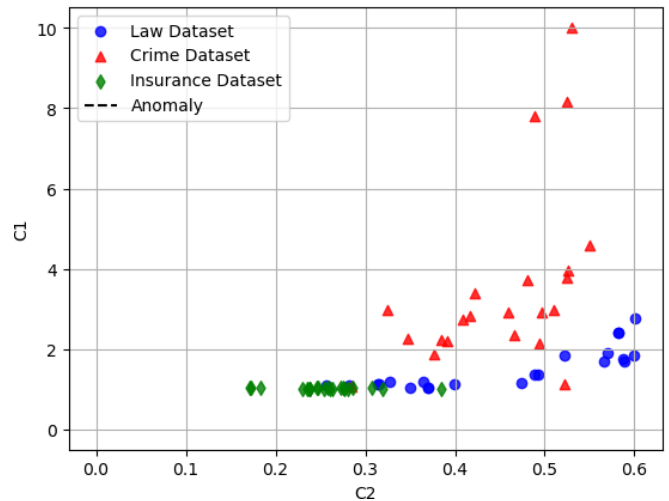

Figure 1: The output of two methods (C1 and C2) approximating the confusion matrix-based metric for various ML models developed on our three datasets, highlighting correlation patterns and anomalies.

Figure 2 illustrates the fairness values derived from each pair of methods approximating parity-based metric for the 22 developed ML models across three datasets. This set of results aligns with the observed Spearman and Pearson correlation results. Notably, except for the P1 and P2 pairs (which exhibit the highest correlation values), most pairs of methods exhibit some degree of irregularity. For example, Figure 2.C compares the output of P1 and P4, and anomalies can be seen in two distinct regions for the Law and Crime datasets. This anomaly is especially notable in the Crime dataset, where these patterns occur repeatedly. The same phenomenon is evident in several other pairs of metrics. For example, in Plot 2.F, a decrease in P2 values in certain cases of the Law dataset corresponds to an increase in P3 values (Other irregularities can be observed in the circled regions of other plots). In summary, our experimental results show that the outcome of methods approximating a specific fairness metric in regression could have a varying level of consistency (depending on the dataset), and even, they could suggest different insights regarding the relative fairness of some ML models.

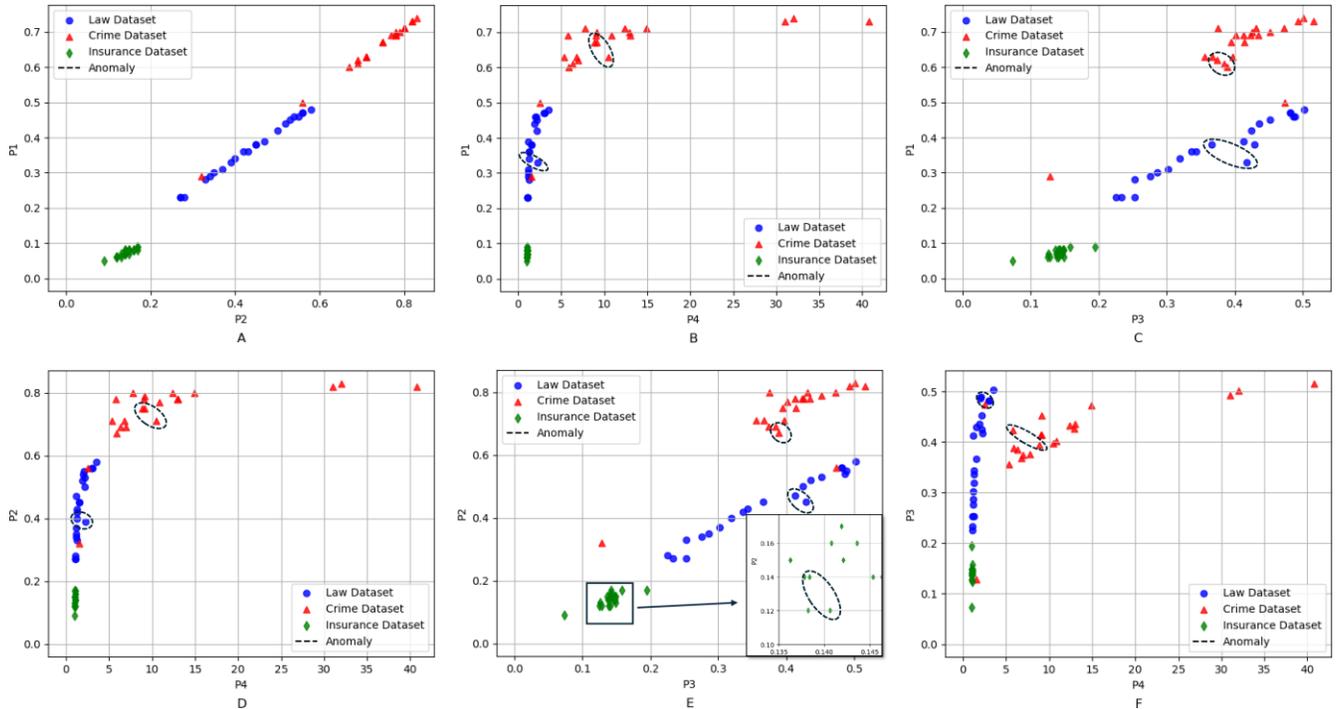

Figure 2: The output of various methods (P1, P2, P3 and P4) approximating the parity-based fairness metric across our three datasets, highlighting correlation patterns and anomalies.

Such inconsistencies could have significant adverse impacts on research area (especially the area of fair regression), and real world. For example, researchers usually compare the effectiveness of their fairness mitigation approaches through examining the amount of improvement in a fairness metric, and based on the result of our research, the choice of methods approximating a specific fairness value could significantly affect our interpretation of increase/decrease in fairness, and the conclusion regarding the effectiveness of various fairness mitigation methods. We recommend that researchers employ multiple fairness measures to enhance the reliability and interpretability of fairness assessments. These challenges also suggest a need for fundamental research to develop more reliable methods for measuring fairness metrics for regression tasks.

## 5 Conclusion and Future Work

This paper studies the consistency of the outcome of various fairness measurement methods in the regression domain. In particular, it focuses on four methods approximating the parity-based fairness metric and two methods estimating the confusion matrix-based fairness metric. Through an extensive set of experiments on various regression tasks, it finds that, the level of consistency between the outcome of various fairness measurement methods could vary significantly. In fact, while the output of certain methods (such as Demographic Parity through Reduction to Classification and Demographic Parity with Wasserstein Barycenters) seem to be strongly aligned with each other across all considered regression tasks, some other pairs of methods, such as (Independence via Probabilistic Classifier-based Density Ratio Estimation, Demographic Parity with Wasserstein Barycenters) or (Separation via Probabilistic Classifier-based Density Ratio Estimation, Equalized Odds) show poor alignment in certain regression tasks, and sometime, they even tend to provide an different insights regarding the relative fairness of two ML models. Therefore, further research needs to be conducted to provide reliable fairness measurement methods for regression tasks.

As a future pathway, we suggest to study the reasons of such anomalies and variability observed across datasets, and comprehensively discover the limitations of various methods. These are major steps towards developing a reliable fairness measurement methods to accurately examine the fairness in the regression domain.